\documentclass[runningheads]{llncs}
\usepackage[T1]{fontenc}
\usepackage{graphicx}
\usepackage{amsmath}
\usepackage{amsfonts}
\usepackage{subcaption}
\usepackage{xcolor}
\usepackage{booktabs}
\usepackage{makecell}
\usepackage{microtype}

\usepackage{url}
\begin{document}
\title{Sustainable techniques to improve Data Quality for training image-based explanatory models for Recommender Systems}
\titlerunning{Sustainable DQ techniques for image-based RS explainability}
%
\author{Jorge Paz-Ruza, David Esteban-Martínez, Amparo Alonso-Betanzos, Bertha Guijarro-Berdiñas}
\authorrunning{J. Paz-Ruza et al.}
%
\institute{Universidade da Coruña, CITIC, Campus de Elviña s/n 15071, A Coruña}
\maketitle              
\begin{abstract}
Visual explanations based on user-uploaded images are an effective and self-contained approach to provide transparency to Recommender Systems (RS), but intrinsic limitations of data used in this explainability paradigm cause existing approaches to use bad quality training data that is highly sparse and suffers from labelling noise. Popular training enrichment approaches like model enlargement or massive data gathering are expensive and environmentally unsustainable, thus we seek to provide better visual explanations to RS aligning with the principles of Responsible AI. In this work, we research the intersection of effective and sustainable training enrichment strategies for visual-based RS explainability models by developing three novel strategies that focus on training Data Quality: 1) selection of reliable negative training examples using Positive-unlabelled Learning, 2) transform-based data augmentation, and 3) text-to-image generative-based data augmentation. The integration of these strategies in three state-of-the-art explainability models increases ($\sim$5\%) the performance in relevant ranking metrics of these visual-based RS explainability models without penalizing their practical long-term sustainability, as tested in multiple real-world restaurant recommendation explanation datasets.
\end{abstract}

\section{Introduction}
\label{chap:introduction}

Recommender Systems (RS) are a pillar of e-commerce, social networks and other fields by offering relevant suggestions to users that boost their engagement and trust \cite{Ricci_Rokach_Shapira_2011}. However, their lack of explainability is also commonplace,  making users sceptical to follow AI-powered recommendations \cite{Zhang_Chen_2020}. Within explainability approaches for RS, visual-based ones see huge popularity \cite{tintarev2015explaining}, as human understanding has visual information as one of its cornerstones. Among the many existing visual explainability approaches, this work is contextualized in the use of user-provided images of items to offer visual explanations of recommendations to other users, which has been explored by different authors in search of more self-contained, relevant and realistic explanations \cite{Díez_Pérez-Núñez_Luaces_Remeseiro_Bahamonde_2020}.

Existing methodologies following this visual explainability approach, such as ELVis \cite{Díez_Pérez-Núñez_Luaces_Remeseiro_Bahamonde_2020}, MF-ELVis \cite{Paz-Ruza_Eiras-Franco_Guijarro-Berdiñas_Alonso-Betanzos_2022} or BRIE \cite{Paz-Ruza_Alonso-Betanzos_Guijarro-Berdiñas_Cancela_Eiras-Franco_2024}, model it as an authorship prediction task: predicting the image of an item that best explains the recommendation of the item to a user is equivalent to predicting the image of the item most likely to be uploaded or authored by that user; this way, the complete task can be modelled as top-N ranking task of user-image authorship. 

The aforementioned existing approaches have progressively improved the handling of existing training data through fixed similarity operators \cite{Paz-Ruza_Eiras-Franco_Guijarro-Berdiñas_Alonso-Betanzos_2022} and reformulation of the training as pairwise ranking \cite{Paz-Ruza_Alonso-Betanzos_Guijarro-Berdiñas_Cancela_Eiras-Franco_2024}, but have not tackled two fundamental data quality limitations of the dyadic user-image data used to achieve explainability: in most contexts, most users have uploaded few to none images to the RS, constituting a \textit{cold start} situation, and data for training the explainability model lacks explicit negative examples, bringing noise into training examples if all unknown user-image pairs are considered cases of bad explanations \cite{Bekker_Davis_2020}.

While these explainability models could be improved with bigger, more expressive topologies or through massive training data gathering efforts, existing literature has proven this approach is not sustainable \cite{Strubell_Ganesh_McCallum_2019}: Energy efficiency and sustainability are key priorities for Machine Learning development, yet these -worryingly popular- approaches carry a unsustainable increase of greenhouse emissions and waste of computational resources. Aligning two different aspects of Responsible AI \cite{dignum2019responsible}, it should be possible to further improve the explainability capabilities of these systems in a computationally sustainable manner. 

In this work, we propose several sustainable techniques to increase the performance of visual RS explanation systems based on user-provided images while respecting their sustainability by focusing on improving the Data Quality (DQ) of the existing training data. The main contributions of this work include:

\begin{itemize}
    \item We design and implement a novel model-agnostic, user-personalized two-step Positive-Unlabelled (PU) Learning model for visual-based RS explainability systems that selects reliable user-image non-authorship relations to act as negative examples within the authorship prediction task, increasing final model performance by refining the label quality of training data. 
    \item We design and implement two data augmentation approaches for visual-based RS explainability systems based on image transforms and text-to-image Generative AI, increasing performance for low-activity users without massive data gathering efforts or model computational overhead. 
    \item We evaluate these techniques by integrating them into explainability models (ELVis, MF-ELVis, BRIE) and testing them against three real-world RS explainability datasets in restaurant recommendation contexts. The computational experiments prove that all three designed strategies not only significantly increase the performance of explanation models in relevant ranking metrics, but also maintain their long-term sustainability in terms of carbon emissions.
\end{itemize}

\section{Background}
\label{chap:background}

This section introduces concepts and formalizations of the tasks and techniques that sustain this research work.

\subsection{Ranking images for visual RS explainability}

Let $\mathcal{U}$ and $\mathcal{I}$ be the sets of users and items that interact in the RS forming $(u, i)$ dyads where $u \in \mathcal{U}$ and $i \in \mathcal{I}$. In the approaches explored in this work, the set $\mathcal{P}$ of images of items taken by users are employed through a reasonable ``authorship'' assumption: if image $p$ of item $i$ could be likely authored by user $u$ in the RS, then image $p$ is a good explanation of a recommendation of item $i$ to user $u$. Therefore, authorship probabilities $Pr(u, p)$ can represent the adequacy of image $p$ to explain to $u$ recommendations of the item shown in $p$, and images of an item can be ranked for their adequacy as explanations using $Pr(u, p)$.

To select from $\mathcal{P}_i$ -the set of images taken of item $i$- the best image $p^*$ to represent a recommendation of $i$ to $u$, we then maximize $Pr(u, p)$ as:

\[
p^* = \arg max_{p \in \mathcal{P}_i} \Pr(u, p).
\]

Trivially, it is true that $Pr(u,p) = 1\;\forall p\in \mathcal{P}_u$ (the set of images taken by user $u$); under authorship modelling, training data does not explicitly contain any negative examples, i.e. bad image explanations such that $Pr(u,p)=0$, constituting a form of Positive and Unlabelled (PU) Data \cite{Bekker_Davis_2020}.

\begin{figure}[htbp]
    \centering
    \includegraphics[width=\linewidth]{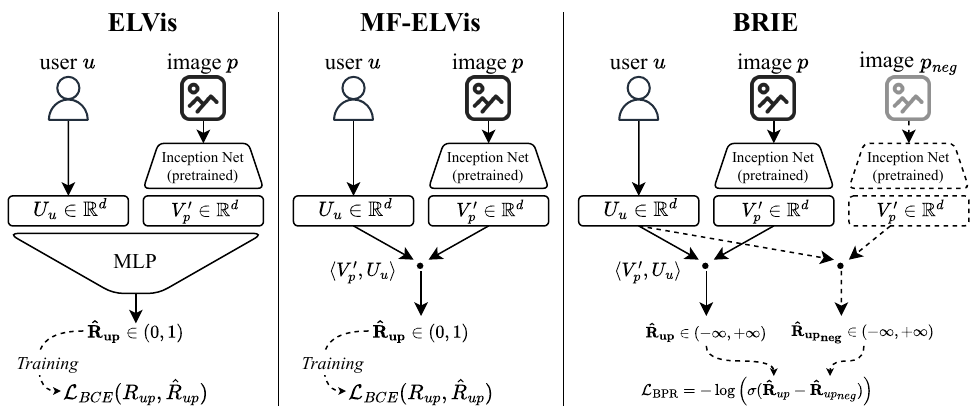}
    \caption{Topologies and optimization of ELVis, MF-ELVis and BRIE \cite{Díez_Pérez-Núñez_Luaces_Remeseiro_Bahamonde_2020,Paz-Ruza_Eiras-Franco_Guijarro-Berdiñas_Alonso-Betanzos_2022,Paz-Ruza_Alonso-Betanzos_Guijarro-Berdiñas_Cancela_Eiras-Franco_2024}.}
    \label{fig:models}
\end{figure}

In the following, we describe existing state-of-the-art approaches ELVis, MF-ELVis and BRIE, that employ this visual-based RS explanation paradigm; their topological and optimization design is shown in Figure \ref{fig:models}.

\subsubsection{ELVis}
Diez et al. \cite{Díez_Pérez-Núñez_Luaces_Remeseiro_Bahamonde_2020} proposed ELVis as a model that uses a neural network-based architecture to compute $Pr(u,p)$. To do this, $u$ (the user id) is mapped into a $d$-dimensional latent embedding, while image $p$ is projected to the same  $d$-dimensional latent space with a \textit{ResNet} image embedding network \cite{Szegedy_Ioffe_Vanhoucke_Alemi_2016} coupled with a dense layer. The two $d$-dimensional vectors, representing $u$'s and $p$'s latent characteristics, are concatenated and projected through a Multi-Layer Perceptron and a final sigmoid activation to obtain as output the authorship probability $Pr(u,p) \in [0,1]$. While ELVis is able to achieve good explanatory performance by modelling users' latent explanatory preferences through implicit feedback, it surrogates training to a binary classification task distant from the real ranking problem, and uses a Deep Learning-oriented architecture that compromises sustainability due to high training and inference computational costs.

\subsubsection{MF-ELVis}

Paz-Ruza et al. \cite{Paz-Ruza_Eiras-Franco_Guijarro-Berdiñas_Alonso-Betanzos_2022} chose to replace ELVIs' MLP architecture by equipping MF-ELVis with fixed embedding similarity function computations (particularly, the inner product $\langle \mathbf{U_u}, \mathbf{V'_p}\rangle$ of $u$'s and $p$'s latent embeddings). MF-ELVis showed results competitive to the more expressive ELVis while reducing training time and emissions owing to a simpler yet effective architecture.

\subsubsection{BRIE}
BRIE \cite{Paz-Ruza_Alonso-Betanzos_Guijarro-Berdiñas_Cancela_Eiras-Franco_2024} seeks to better handle existing training data by modelling learning as a pairwise ranking task \cite{Rendle_Freudenthaler_Gantner_Schmidt-Thieme_2012} and resampling the paired random negative examples (unrelated image-user pairs) at the start of every epoch; BRIE is the current state-on-the-art in image ranking-based recommendation explainability and provides both significantly higher performance and lower training and inference execution time and greenhouse emissions than ELVis and MF-ELVis.

\subsection{Data Quality on dyadic data}

While BRIE and other models improve the handling of training data, they do not tackle some intrinsic quality issues of dyadic data: the \textit{cold start} situation affecting most users, which do not upload many images into the system, and the unlabelled nature of the assumed negative examples.

\subsubsection{Positive and Unlabelled Learning}

Existing models assume any unrelated $(u,p)$ pair can be considered a negative example (bad explanation) which is incompatible with images of a user being useful explanations for other users; ultimately, this produces noisy training and worse explanations. 

In PU Data contexts, PU Learning is a semi-supervised learning paradigm that directly handles PU Data to improve training and model performance \cite{Bekker_Davis_2020}. We focus on the most popular and flexible family of methods, named \textit{two-step approaches}. Let $P$ be the set of known positive examples (here, the set of past image uploads by users in the RS ($\mathcal{D}=\{(u, i, p),\dots\}$) and $U$ the set of unlabelled examples (here, all other user-item-image combinations not in the original set ($\{(u, i, p),\dots\}\;\; p\in \mathcal{P} \setminus \mathcal{P}_u$), which considers both positive examples (good explanations) and negative examples (bad explanations). First, two-step PU Learning approaches obtain a refined set $RN \subset U$ of ``reliable'' negative examples (in this work, images that are, with certainty, bad explanations for a given $u$). Then, as the second step, a final model is trained using $P$ and $RN$ to obtain a more accurate model trained with minimal noise; moreover, this refinement in the quality of training data also allows for shorter, more efficient model training with smaller and higher quality training sets (because  $|RN| < |U|$).

\subsubsection{Data Augmentation}

Data Augmentation is a popular approach to increase the availability of coherent training data without massive, costly data gathering  \cite{van2001art}, increasing the overall quality of existing data. We are interested in using data augmentation in some of its image variants to tackle the \textit{cold start} problem in our task, where users have uploaded (typically few) images of items to the RS:

\begin{itemize}
    \item \textbf{Transform-based augmentation}, which uses existing examples to introduce new coherent data in the domain, such as with geometric, contrast, deformations or cutout transforms \cite{shorten2019survey}; these can replicate the variety and imperfections of user-uploaded images in RS. 
    \item \textbf{Generative augmentation}, which can directly create new training examples; modal transform models, such as text-to-image \cite{yin2023ttida}, could potentially use existing textual data in the RS (e.g. users' reviews) to enrich training.
\end{itemize}

\section{Proposed sustainable Data Quality (DQ)  techniques}
\label{chap:tecnicas}

This section covers the three novel proposed methods (based on PU Learning, transform-based data augmentation, and generative data augmentation) to sustainably increase the performance of visual-based explainability of RS based on user-provided images through improvements in the quality of training data. 

\subsection{Reducing label noise with PU Learning}

To overcome the labelling noise caused by the naive negativity assumptions on the task's PU Data made by existing approaches, we propose to apply PU Learning to properly handle unseen (user, image) pairs as unlabelled data.

One challenge compared to classic PU Learning is the complexity of dyadic user-image data that forms the unlabelled set $U$. For instance, the same image uploaded by user $u$ is unlabelled for users $u'$ and $u''$ but can be respectively a good and bad explanation for those users. To overcome this limitation we model PU Learning user-wise, solving the PU task independently for each user; we choose to follow the two-step paradigm \cite{Bekker_Davis_2020} for its flexibility. This choice is held by two basic assumptions: smoothness of the positive class (for each user, good explanations are images similar to those from the user), and separability with the negative class (for each user, bad explanations are images that differ from those from the user). We also assume SCAR (Selected Completely at Random) as the labelling mechanism \cite{Bekker_Davis_2020}, as it is reasonable to consider the user's own images are an i.i.d. sample of the distribution of good explanations for them. 

To select reliable negative examples for each user we used discriminators based on a Rocchio classifier, which is light-weight and popular in classic PU Learning \cite{rocchio71relevance}. The selection of user-wise reliable negatives $RN_u$ of each user is illustrated in Figure \ref{fig:pu-learning} and proceeds as follows: 

\begin{figure}
    \centering
    \includegraphics[width=0.9\linewidth]{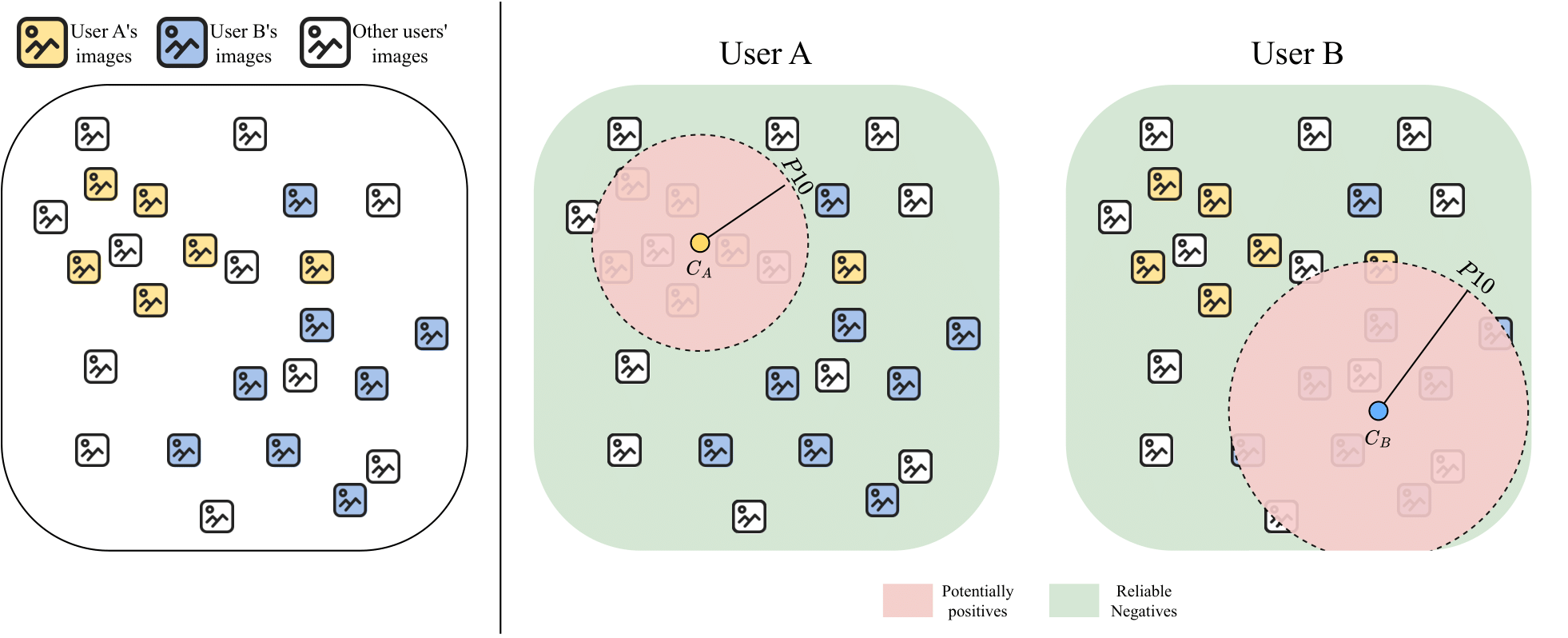}
    \caption{User-personalized PU Learning technique proposed to select reliable negative examples (bad image explanations) for each user in recommendation personalization contexts. Here, the decision boundary for reliable negative selection is shown for two users A and B with different explanatory preferences.}
    \label{fig:pu-learning}
\end{figure}

\begin{enumerate}
    \item Find the centroid of the embedding projections of $u$'s images as obtained by an image embedding model $\phi$ to obtain a prototype of a good image explanation to $u$ as $C_u = \frac{1}{|\mathcal{P}_u|} \sum_{p \in \mathcal{P}_u} \phi(p)$.

    \item Discriminate reliable negatives through a similarity threshold; we employ the 10th percentile $P_{10}$ of the cosine similarities $SC$ between $u$'s images and its centroid $C_u$ to filter outliers in $\mathcal{P}_u$.
    \item Admit unlabelled user-image pairs $(u, p)$ with $p\notin\mathcal{P}_u$ to $RN_u$ as reliable negative examples (i.e. $p$ is a bad explanatory image for $u$ because they would not author this image) if $SC(C_u, \phi({p})) \leq P_{10}$.
\end{enumerate}

Finally, the explanation model can be seamlessly trained using the known positive examples $P$ (i.e. $\mathcal{D}$, the user-image upload tuples originally in the data, representing good explanations) and the union of the reliable negative example sets $RN_u$ (i.e. user-image pairs where the user would be very unlikely upload such an image, representing bad explanations). Since $RN$ is a refined, smaller set of $U$, the consequent lower labelling noise  and increased quality of training data leads to more effective and efficient training of explanatory models.

\subsection{Data Augmentation}

To tackle the \textit{cold start} problem (inability to learn from inactive users) due to the data sparsity in this task, we propose two Data Augmentation methods to enrich positive examples and increase training data quality for low-activity users by using either image transforms or text-to-image example generation.

\subsubsection{Transform-based augmentation}
\label{sec:data-aug}

\begin{figure}[h]
    \centering
    \begin{subfigure}{0.24\textwidth}
        \centering
        \includegraphics[width=\textwidth]{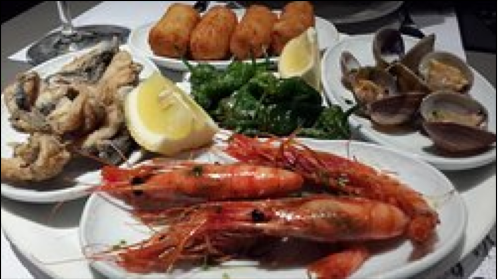}
        \caption{Original}
    \end{subfigure}
    \begin{subfigure}{0.24\textwidth}
        \centering
        \includegraphics[width=\textwidth]{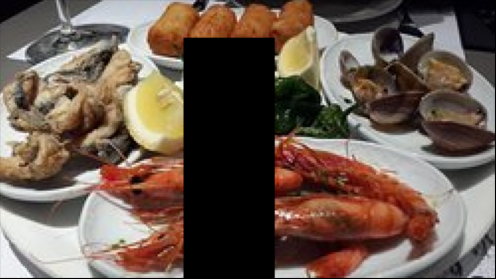}
        \caption{Random cutout}
    \end{subfigure}
    \begin{subfigure}{0.24\textwidth}
        \centering
        \includegraphics[width=\textwidth]{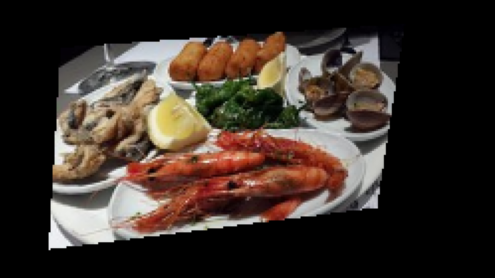}
        \caption{Deformation}
    \end{subfigure}
    \begin{subfigure}{0.24\textwidth}
        \centering
        \includegraphics[width=\textwidth]{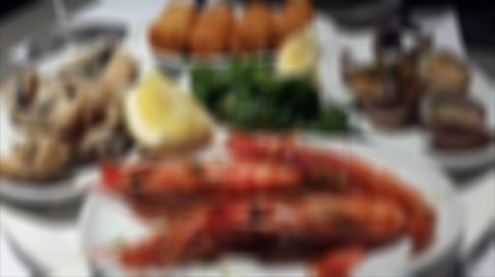}
        \caption{Gaussian noise}
    \end{subfigure}

    \caption{Example of transform-based data augmentation based on user's existing images, here to an image uploaded to a restaurant recommendation review.}
    \label{fig:data_augmentation}
\end{figure}

Following existing literature, we employed three advanced transforms to user images in the RS: random cutouts, geometric transforms and Gaussian blur, mimicking the imperfections in user-uploaded images, as exemplified in Figure \ref{fig:data_augmentation}. 

Augmentation is most effective for low-activity users, so we set a user minimal activity threshold. For each user with $|P_u|<n$ images, we generate random transforms $p'$ of images $p \in P_u$ and add them to $P_u$ until reaching $|P_u|=n$, enriching in a personalized manner the existing data for each low-activity user. 

\subsubsection{Text-to-image generative data augmentation}
\label{sec:gen-ai}

We also propose the use of generative models to directly create new good explanation image data for less active users to enrich their training data. User reviews of items typically include a textual review representative of user preferences, so we employ a text-to-image Generative AI model to leverage the existing textual information in the RS. 

For each user $u$ with $|P_u|<n$ reviews, we consider the set of textual reviews $R_u$ formed by reviews $T$ written by user $u$. Using a generative text-to-image model $G$, and until an activity threshold $|P_u|=n$ is reached, we iteratively generate images $p_{gen} = G(T)$ for user $u$ from randomly selected reviews $T \in R_u$ using the prompt exemplified in Figure \ref{fig:amusedexample}:  

\begin{figure}[h]
    \centering
    \begin{minipage}{0.4\textwidth}
        \textit{Photorealistic image, taken with a smartphone camera, uploaded with the following} \textbf{\textcolor{blue}{<type of item>}} \textit{review}: \textbf{\textcolor{blue}{<review text>}}
    \end{minipage}
    \hfill
    \begin{minipage}{0.55\textwidth}
        \centering
        \includegraphics[width=.235\textwidth]{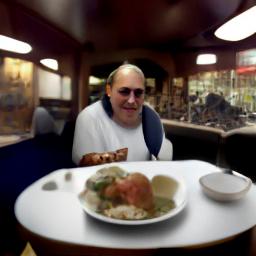} 
        \includegraphics[width=.235\textwidth]{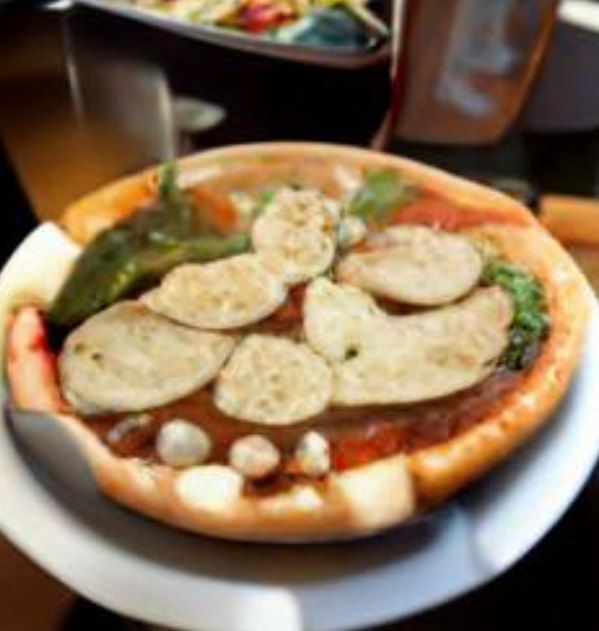} 
        \includegraphics[width=.235\textwidth]{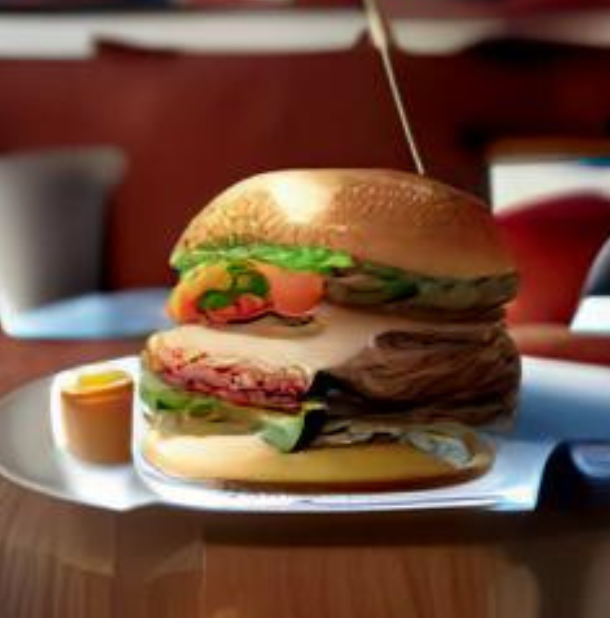} 
        \includegraphics[width=.235\textwidth]{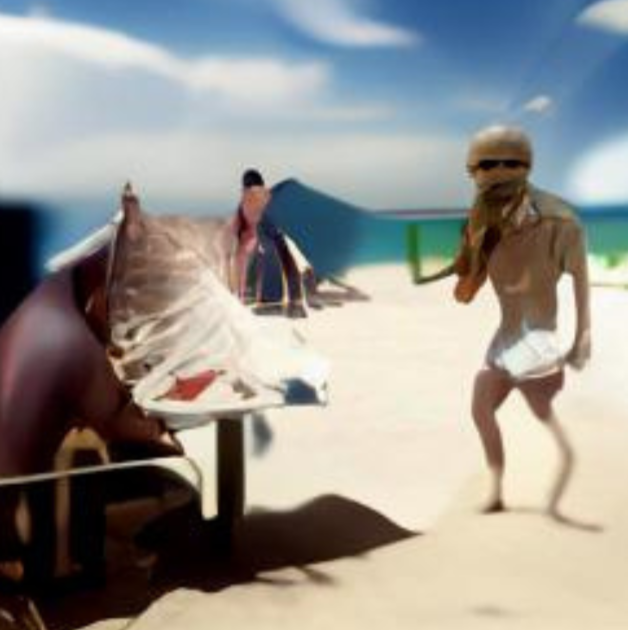} 
    \end{minipage}%
    \caption{Prompt structure (left) and examples of generated training images from reviews (right) through generative augmentation in a restaurant context.}
    \label{fig:amusedexample}
\end{figure}

\section{Experimental Setup}

This section covers different aspects of the experimental configurations including datasets, evaluation protocols and other implementation details.

\subsection{Datasets}

We evaluated our techniques using real-world datasets in image-based restaurant recommendation explanation in TripAdvisor \cite{pablo_perez_nunez_2025_14622324}, following existing approaches, in three different cities (Gijon, Madrid, and Barcelona). We employed the user-wise leave-one-out dataset partitions provided by Díez et al. \cite{Díez_Pérez-Núñez_Luaces_Remeseiro_Bahamonde_2020}: for each user with $N\geq2$ images, one is used for training and the $N-1$ for testing; in the test set, for each $(u,i,p)$ positive tuple (assumed good explanation), all $(u,i,p')$ with $p' \in P_i \setminus \{p\}$ tuples are added as negative examples to conform each image ranking test case. This is, we assume an offline evaluation \cite{Shani_Gunawardana_2011} where the system performs better the higher it ranks for a user $u$ its real image $p$ of item $i$, where $p$ was not seen during training by the model. Table \ref{tab:city_summary} shows the basic statistics of each of the datasets employed to evaluate our techniques. 
\begin{table}[htbp]
\caption{Summary of TripAdvisor datasets on image-based recommendation explanation in restaurant contexts.}
\label{tab:city_summary}
\centering
\resizebox{.5\textwidth}{!}{
\begin{tabular}{ccccc}
\toprule
\textbf{Dataset} & \textbf{Users} & \textbf{Restaurants} & \textbf{Images} & \textbf{Images/User} \\
\midrule
Gijón & 5,139 & 598 & 18,679 & 3.64 \\
Barcelona & 33,537 & 5,881 & 150,416 & 4.49 \\
Madrid & 43,628 & 6,810 & 203,905 & 4.67 \\
\bottomrule
\end{tabular}
}

\end{table}
\subsection{Evaluation}

With respect to explanation ranking quality, we followed recent works \cite{Paz-Ruza_Alonso-Betanzos_Guijarro-Berdiñas_Cancela_Eiras-Franco_2024}, and evaluated Recall@k and NDCG@k with $k=10$, and AUC-ROC. Recall@10 and NDCG@10 only considered test cases with $>10$ images, as trivially Recall@10=1 and NDCG@k>0 otherwise. Previous works also restrict evaluation to users with $>10$ train images (active users), but we evaluate using all users as our techniques focus on improving training quality for low-activity users.

With respect to model sustainability, we analyze greenhouse emissions and execution times of the models and techniques, using CodeCarbon \cite{benoit_courty_2024_11171501}, by measuring the computational cost of applying our techniques to the datasets and training existing models on the original and refined datasets. Note that all techniques only affect the training dataset and consequent model training, so models' inference emissions and execution time are not affected by them. 
\subsubsection{Implementation details}

We now describe some finer implementation details regarding model configurations and experimental setup: 

\begin{itemize}
    \item Regarding model hyper-parameters, we follow the respective best hyperparameter settings reported by the authors of each model and only re-optimize the number of training epochs used on the datasets refined with our techniques, using the same early-stopping policy with $p=5$ and $\Delta=0.001$ against the validation set of each dataset. 
    \item We use aMuseD256 \cite{Patil_Berman_Rombach_Platen_2024} for text-to-image in generative-based data augmentation, and ViT Large 14 \footnote{\url{https://huggingface.co/openai/clip-vit-large-patch14}} to obtain image embeddings.
    \item Our improvement techniques for image-based RS explainability are made available in an open-source Python library\footnote{\url{https://pypi.org/project/data-improvement-library/}} that is model-agnostic and modular to allow its research and commercial use.
\end{itemize}

\section{Results}

This section covers the results of performance and sustainability experiments with our techniques for improving image-based recommender explainability.

\subsection{Explanation ranking performance}

Table \ref{tab:my-table} shows the explanation ranking qualities of ELVis, MF-ELVis and BRIE with and without applying our proposed techniques. For most datasets and metrics, the best results are obtained by applying our proposed techniques to the training data, with relative improvements up to 5\% compared to not applying them.

Techniques do not clearly outperform others despite their diversity, but all increase performance in a majority of dataset and model combinations. Using our techniques, ELVis surpasses the state-of-the-art model BRIE, which shows little benefit from using them; BRIE employs a strong training regularization which may dilute the effect of our techniques \cite{Paz-Ruza_Alonso-Betanzos_Guijarro-Berdiñas_Cancela_Eiras-Franco_2024}.

\begin{table}[]
\centering
\caption{Ranking quality results of ELVis, MF-ELVis and BRIE with and without applying our PU Learning (PU), transform-based data augmentation (T-DA) and generative data augmentation (Gen-DA) techniques, tested on TripAdvisor datasets. The best result for each dataset-metric combination is bolded and results with daggers † are significantly better than their no-technique counterparts.}
\label{tab:my-table}
\resizebox{\textwidth}{!}{%
\begin{tabular}{@{}llllllllllll@{}}
\toprule
                  & \multicolumn{3}{l}{\textbf{Gijon}}                                                                                                                                                       &  & \multicolumn{3}{l}{\textbf{Barcelona}}                                                                                                                                                                       &  & \multicolumn{3}{l}{\textbf{Madrid}}                                                                                                                                                      \\ \cmidrule(lr){2-4} \cmidrule(lr){6-8} \cmidrule(l){10-12} 
                  & \textbf{\begin{tabular}[c]{@{}l@{}}Recall\\ @10\end{tabular}} & \textbf{\begin{tabular}[c]{@{}l@{}}NDCG\\ @10\end{tabular}} & \textbf{\begin{tabular}[c]{@{}l@{}}AUC\\ ROC\end{tabular}} &  & \textbf{\begin{tabular}[c]{@{}l@{}}Recall\\ @10\end{tabular}} & \textbf{\begin{tabular}[c]{@{}l@{}}NDCG\\ @10\end{tabular}} & \multicolumn{1}{c}{\textbf{\begin{tabular}[c]{@{}c@{}}AUC\\ ROC\end{tabular}}} &  & \textbf{\begin{tabular}[c]{@{}l@{}}Recall\\ @10\end{tabular}} & \textbf{\begin{tabular}[c]{@{}l@{}}NDCG\\ @10\end{tabular}} & \textbf{\begin{tabular}[c]{@{}l@{}}AUC\\ ROC\end{tabular}} \\ \midrule
ELVis             & 0.492                                                         & 0.262                                                       & 0.702                                                      &  & 0.562                                                         & 0.320                                                       & 0.726                                                                          &  & 0.522                                                         & 0.298                                                       & 0.737                                                      \\
MF-ELVis          & 0.486                                                         & 0.278                                                       & 0.691                                                      &  & 0.527                                                         & 0.304                                                       & 0.696                                                                          &  & 0.490                                                         & 0.281                                                       & 0.699                                                      \\
BRIE              & \textbf{0.535}                                                & \textbf{0.306}                                                       & 0.736                                                      &  & 0.584                                                         & 0.342                                                       & 0.745                                                                          &  & 0.538                                                         & \textbf{0.316}                                              & 0.752                                                      \\ \midrule
Elvis + PU        & 0.517†                                                        & 0.280†                                                      & 0.715†                                                     &  & \textbf{0.587†}                                               & 0.334†                                                      & 0.747†                                                                         &  & 0.524†                                                        & 0.307†                                                      & 0.752†                                                     \\
ELVis + T-DA      & 0.532†                                                        & 0.290†                                                      & 0.720†                                                     &  & \textbf{0.587†}                                               & 0.336†                                                      & 0.745†                                                                         &  & 0.531†                                                        & 0.315†                                                      & 0.752†                                                     \\
ELVis + Gen-DA    & 0.527†                                                        & 0.292†                                                      & 0.722†                                                     &  & 0.582†                                                        & \textbf{0.344†}                                             & \textbf{0.750†}                                                                &  & 0.534†                                                        & 0.309†                                                      & \textbf{0.755†}                                            \\
MF-Elvis + PU     & 0.490                                                         & 0.276                                                       & 0.692†                                                     &  & 0.558†                                                        & 0.315†                                                      & 0.702†                                                                         &  & 0.499†                                                        & 0.295†                                                      & 0.716†                                                     \\
MF-ELVis + T-DA   & 0.495†                                                        & 0.280†                                                      & 0.699†                                                     &  & 0.527                                                         & 0.302                                                       & 0.688                                                                          &  & 0.500†                                                        & 0.287†                                                      & 0.709†                                                     \\
MF-ELVis + Gen-DA & 0.492                                                         & 0.284†                                                      & 0.705†                                                     &  & 0.526                                                         & 0.302                                                       & 0.693                                                                          &  & 0.512†                                                        & 0.298†                                                      & 0.713†                                                     \\
BRIE + PU         & 0.531                                                         & 0.299                                                       & 0.731                                                      &  & 0.586†                                                        & 0.342                                                       & 0.746†                                                                         &  & 0.538                                                         & 0.314                                                       & 0.753†                                                     \\
BRIE + T-DA       & 0.529                                                         & 0.301                                                       & 0.728                                                      &  & 0.584                                                         & \textbf{0.344†}                                             & 0.745                                                                          &  & 0.535                                                         & 0.312                                                       & 0.748                                                      \\
BRIE + Gen-DA     & \textbf{0.535}                                                & \textbf{0.306}                                              & \textbf{0.737†}                                            &  & 0.583                                                         & 0.340                                                       & 0.745                                                                          &  & \textbf{0.540†}                                               & 0.315                                                       & 0.752                                                      \\ \bottomrule
\end{tabular}%
}
\end{table}

\subsection{Impact on model sustainability}

\begin{figure}[htbp]
    \begin{minipage}[][][t]{.5\textwidth}
        
\centering

\captionof{table}{Training carbon emissions and execution time for explainability models ELVis, MF-ELVis and BRIE with and without the proposed Data Quality explainability techniques.}
\label{tab:emissions}
\resizebox{\textwidth}{!}{%
\begin{tabular}{@{}lll@{}}
\toprule
                  & \textbf{\makecell{Time\\ (hh m' s")}} & \textbf{\makecell{Emissions \\(gCO$_2$e)}} \\ \midrule
ELVis             & 62' 11"                  & 7.96                          \\
MF-ELVis          & 13' 24"                  & 2.67                          \\
BRIE              & 12' 33"                  & 1.49                          \\ \midrule
ELVis + PU        & 22' 53"†                 & 4.86†                         \\
ELVis + T-DA      & 36' 26"†                 & 7.84†                         \\
ELVis + Gen-DA    & 5h 20' 10"               & 57.58                         \\
MF-Elvis + PU     & \textbf{8' 32"†}         & \textbf{1.17†}                \\
MF-ELVis + T-DA   & 30' 12"                  & 6.10                          \\
MF-ELVis + Gen-DA & 3h 59' 28"               & 47.12                         \\
BRIE + PU         & 15' 31"                  & 1.73                          \\
BRIE + T-DA       & 29' 25"                  & 4.92                          \\
BRIE + Gen-DA     & 4h 03' 43"               & 47.71                         \\ \bottomrule
\end{tabular}%
}
\end{minipage}
\begin{minipage}[][][t]{.5\textwidth}
    \centering
    \includegraphics[width=\linewidth]{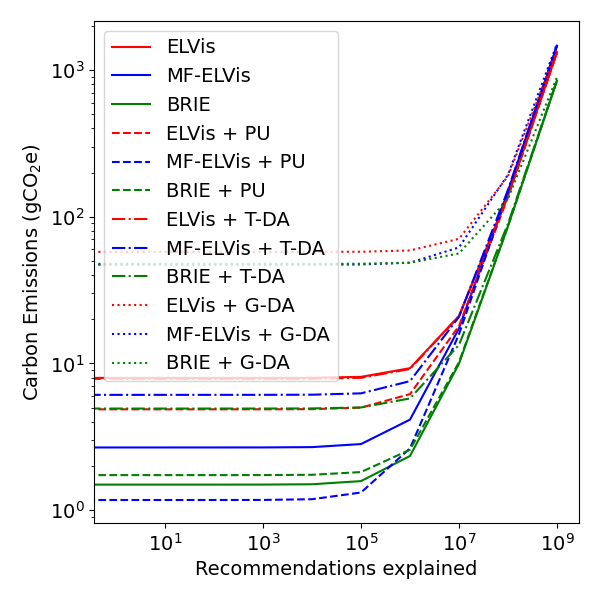}
    \caption{Total carbon emissions (training and inference) depending on the number of inference cases for explainability methods ELVis, MF-ELVIs and BRIE with and without our techniques for Data Quality improvement.}
    \label{fig:enter-label}
\end{minipage}
\end{figure}

Table \ref{tab:emissions} shows the training carbon emissions and execution time of the analyzed models with and without applying our techniques for DQ improvement. Our PU-based technique allows the model to perform an up to 60\% shorter, more effective training on refined and smaller training datasets with less label noise. Data augmentation approaches also increase model performance, but the larger training set and example generation carry higher computational overheads.

The long-term inference use of explainability models affects sustainability more than their training (e.g. TripAdvisor receives 100M reviews per year \cite{cycles_Text}), and our techniques do not have effects on model inference. Figure \ref{fig:enter-label} shows the long-term emissions (training+inference) of ELVis, MF-ELVis and BRIE with and without our techniques; clearly, the possible computational overheads of using data quality to improve model performance become negligible compared to inference-time emissions.

\section{Conclusions}

This work is framed at the intersection of three of the main modern challenges in Recommender Systems: explainability, sustainability and personalization. In the context of visual-based RS explainability through user-provided item images, we proposed to sustainably improve the performance of explainer models' performance by focusing on increasing Data Quality of the training data.

We proposed three novel techniques: one based in PU Learning, to tackle label noise arising from naive assumptions on data, and two based in Data Augmentation, either transform-based or text-to-image generation-based, to tackle \textit{cold start} problems common to most users. Our computational experiments in real-world restaurant recommendation explanation tasks show that these techniques increase performance ($+\sim5\%$)  of existing models (ELVis, MF-ELVis and BRIE) without compromising their long-term sustainability.

With respect to further work, we identify different open ends for future experimentation: 1) a proper hyper-parameter re-optimization of the used explainability models which may the effectiveness of our techniques; 2) the use of datasets that contain image-less textual reviews, that could further prove the effects of generation-base augmentation 3) the design of PU Learning assumptions beyond single-prototype characterizations of users, which may be needed for users with diverse explanatory tastes.

\begin{credits}

\subsubsection{\discintname}
The authors have no competing interests to declare that are
relevant to the content of this article.
\end{credits}
%
%
%
\bibliographystyle{splncs04}
\bibliography{EXPDQ}

\end{document}